# DEPLOYMENT OF MOBILE ROUTERS ENSURING COVERAGE AND CONNECTIVITY

Emi Mathews[1] and Ciby Mathew[2]

[1]Heinz Nixdorf Institute, University of Paderborn, Germany
emi@hni.upb.de
[2]Department of Computer Science, University of Paderborn, Germany
ciby@mail.upb.de

## ABSTRACT

*Maintaining connectivity among a group of autonomous agents exploring an area is very important, as it promotes cooperation between the agents and also helps message exchanges which are very critical for their mission. Creating an underlying Ad-hoc Mobile Router Network (AMRoNet) using simple robotic routers is an approach that facilitates communication between the agents without restricting their movements. We address the following question in our paper: How to create an AMRoNet with local information and with minimum number of routers? We propose two new localized and distributed algorithms 1) agent-assisted router deployment and 2) a self-spreading for creating AMRoNet. The algorithms use a greedy deployment strategy for deploying routers effectively into the area maximizing coverage and a triangular deployment strategy to connect different connected component of routers from different base stations. Empirical analysis shows that the proposed algorithms are the two best localized approaches to create AMRoNets.*

## KEYWORDS

*Ad-hoc Mobile Router Network, Coverage, Connectivity, Localized deployment*

## 1. INTRODUCTION

We envision a scenario with several agents which are humans or powerful robots moving autonomously on a terrain represented by a plane. These autonomous agents begin their exploration from one or more stationary base camp(s). We are looking for local and distributed strategies for maintaining the connectivity of the agents with the base station(s) and the other agents, as it promotes cooperation between the agents and also helps message exchanges which are very critical for their mission. These strategies must not restrict agent movements for the sake of maintaining connectivity.

Scenarios such as urban search and rescue and exploration of an unknown terrain are good examples, where we often have several exploring agents and one or more base station(s). In urban search and rescue scenarios, due to the aftermath of natural or manmade disasters such as earthquakes, tsunamis, hurricanes, wars or explosions, the fixed communication infrastructure that could support communication between the agents are often destroyed. In other scenarios such as exploration of unknown terrains, e.g. subterranea or remote planets, no such infrastructure to support communication exist. A line of sight communication between the agents is not possible in such complex scenarios as the distance between the agents are often very large due to the large area to be explored. The presence of obstacles makes it difficult even at shorter inter-agent distance.

A stable and high bandwidth communication is feasible if we employ a multi-hop ad-hoc networking strategy for the agents. However, in such scenarios the number of agents is often very limited. Hence the agents themselves could not form a connected network always. Moreover, if they try to keep the network connected, it would restrict their movements.

We propose an alternate solution to maintain connectivity of the agents, use swarm robots to create a network that acts as an infrastructure to support the communication of the agents. Swarm robots are simple and low cost robots, usually available in large numbers, perform complex behaviors at the macro-level, with high level of fault tolerance and scalability. The behaviours emerge from simple local interactions between individual robots.

The swarm robots act as routers for agents' communication. Thus we have a two tier network, with the agents and base stations lying at the upper layer and the routers deployed at the lower layer. The lower layer created to facilitate the communication between upper layer members is called Ad-hoc Mobile Router Network (AMRoNet). This network, in addition to supporting upper layer members' communication, provides various services to the agents, such as location information, topological maps and shortest path to base stations, and can also assist the search and rescue operation of the agents. The main advantage of this network is that the routers could relocate and maintain the connectivity in case of failures which are very common in scenarios described above.

In this paper, we address the following question: How to create an AMRoNet with local rules and with minimum number of routers? The remainder of this paper is organized as follows: Section 2 introduces the scenario and notations used in this paper, formalizes the problem and provides a brief overview on the theoretical background of the problem. Related approaches known from the literature are discussed in Section 3. In Section 4 we present two new algorithms for creating AMRoNet with local information. Next, in Section 5 we present a simulation based experimental evaluation and in Section 6 the analysis of the proposed algorithms. Finally, Section 7 summarizes the main results of this work and provides an outlook on possible future research.

This paper is partly an extension of the contribution published in the proceedings of NetCom 2012 [16]. However, in this paper one of the algorithms for creating AMRoNet and its experimental evaluation are completely new with respect to [16].

## 2. PRELIMINARIES

### 2.1. Basic Assumptions

We have a two tier network, with the agents and base stations forming the upper layer. The environment where the agents explore is a 2-D area denoted as $A$ and has $n$ base stations. There are $N_a$ agents which are humans or robots capable of performing tasks such as urban search and rescue. As the focus of this paper is mainly on the AMRoNet, we do not specify the requirements of the agents and the base stations, which vary according to the scenario considered. The only assumption we make is that they have wireless devices to support communication.

The lower layer forming the AMRoNet consists of total $N_r$ routers. The routers denoted by $R$ are very simple robots compared to the agents with limited sensing capabilities with which they avoid obstacles and perform local navigation. Routers are equipped with wireless transceivers for communication.

We assume the unit disk graph wireless model [3] for communication, where each node (agent, router or base station) can communicate with others located within a circle of radius $r_c$. We also assume that the communication area of one node is much less than *A*. Hence, the agents have to send packets over several routers to reach a particular destination (other agent or base station).

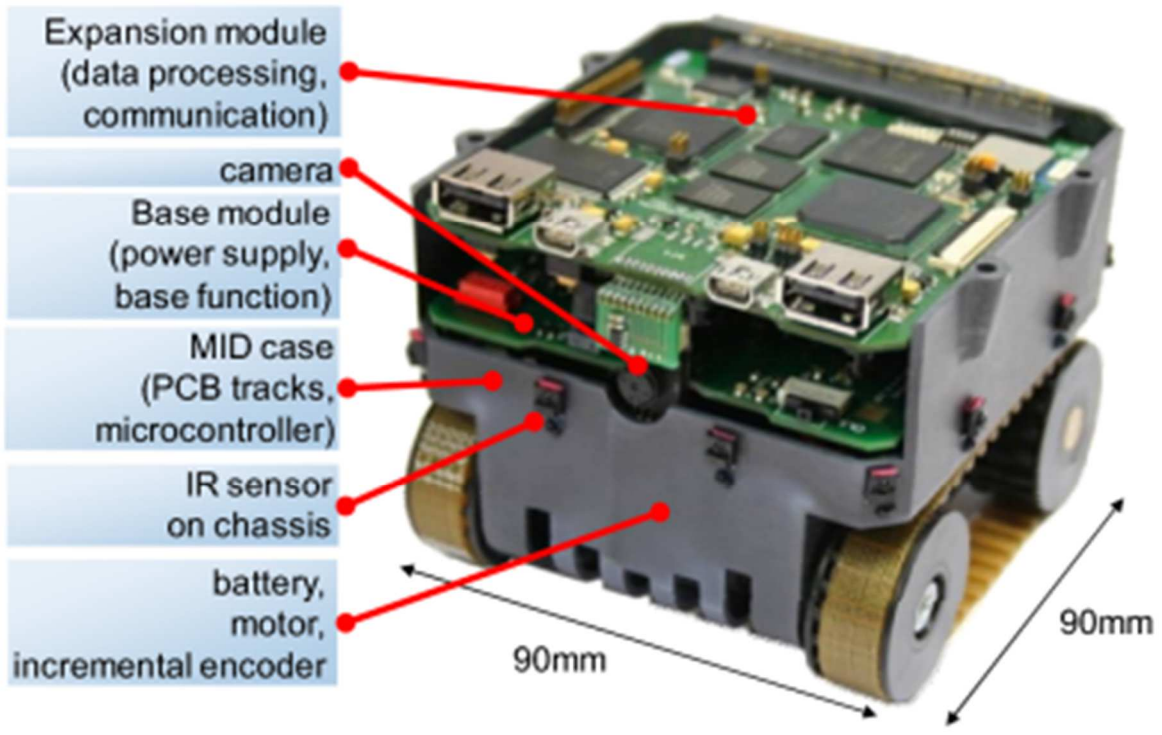

**Figure 1: Bebot mini-robot**

Mini-robots such as Bebots [8], shown in Figure 1 are suitable candidates for routers. These robots are equipped with a camera with which they can assist agents in search and rescue operations. They have an infrared sensor ring for obstacle avoidance and wifi, zigbee and bluetooth modules for communication.

## 2.2. Objective

We are interested in maintaining connectivity of the agents with minimum number of routers. Hence our objective is to find a strategy to create AMRoNet that maximizes the total communication area covered. Let $\pi r_c^2$ be the communication area of one robot which is denoted as $A_i$. So our objective is to maximize *coverage* which is defined as

$$Coverage = \frac{\cup_{i=1}^{N} A_i}{A} \quad (1)$$

## 2.3. Optimal Deployment

We can find the optimal router location of an AMRoNet from the static optimal placement strategies used in the area coverage problems. The objective of these problems is to place minimum number of nodes in an environment such that, every point is optimally covered. If we look at the *optimal coverage* with respect to the total sensing area, the robots could form a triangular grid as shown in Figure 2(a). When the inter-robot distance $d = \sqrt{3}.\, r_s$, where $r_s$ is the sensing radius, 100% coverage is attained with minimum number of robots. This approach creates a connected network if $\frac{r_c}{r_s} \geq \sqrt{3}$.

However, our interest is on the *communication area coverage*. A triangular grid with the inter-robot distance $d = \sqrt{3}.\, r_c$ cannot provide 100% communication area coverage, as robots cannot communicate when $d > r_c$. So a coverage and connectivity (*C–C*) constraint arises and our objective is to maximize the communication area coverage with connectivity.

If we create a triangular grid with reduced inter node distance $d = r_c$, it is not optimal according to the *C–C* constraint. What is optimal in 1-D, is an r-strip shown at the bottom row of Figure 2(b), where $d = r_c$. In 2-D, the lower bound on node density for optimal *C–C* is $d_{OPT} \geq \frac{0.522}{r^2}$ [12]. The optimal solution that achieves communication coverage with 1-connectivity in 2-D is the r-strip tile shown in the Figure 2(b) [1]. It has a spatial density $d_{STR} = \frac{0.536}{r^2}$ [12]. The r-strip tile in 2-D is created as follows: for every integer $k$ place a strip horizontally such that there is one node positioned at $\left(0, k\left(\frac{\sqrt{3}}{2}+1\right)r_c\right)$ for every even $k$, and one node positioned at $\left(\frac{r_c}{2}, k\left(\frac{\sqrt{3}}{2}+1\right)r_c\right)$ for every odd k. Finally place some nodes vertically in the following way. For every odd $k$, place two nodes at $\left(0, k\left(\frac{\sqrt{3}}{2}+1\right)r_c \pm \frac{\sqrt{3}}{2}\right)$. The purpose of this vertical strip is to connect the horizontal strips and thus ensure connectivity between all nodes.

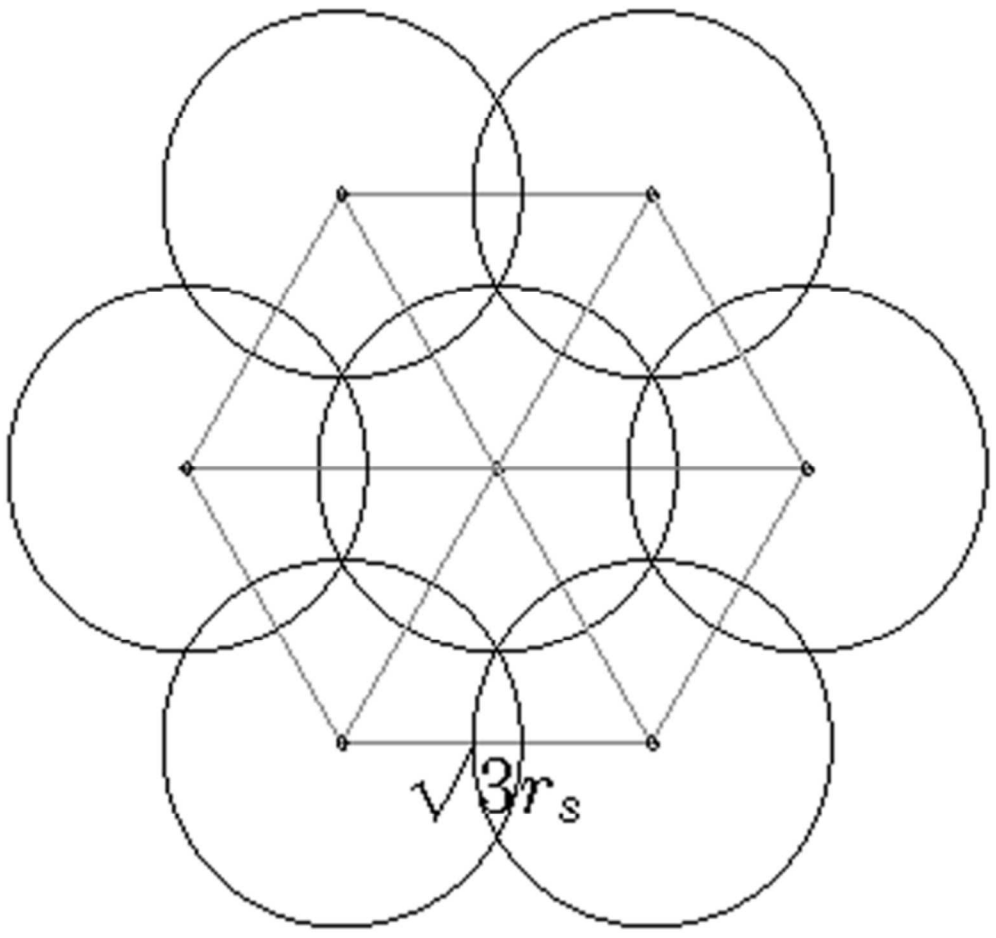

(a) Optimal Coverage

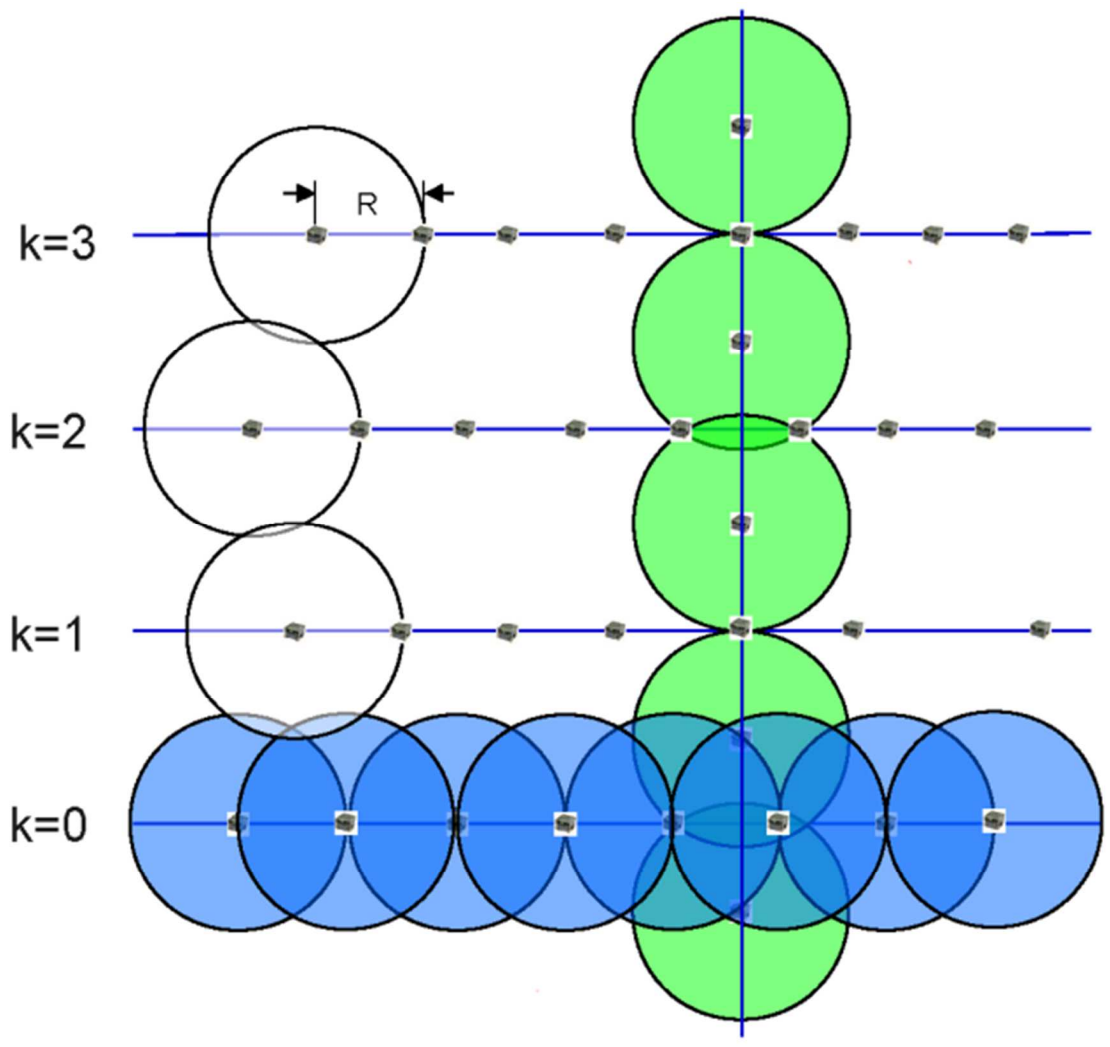

(b) R-strip tile in 2D

Figure 2: Coverage and Connectivity

More commonly used regular patterns are hexagonal grid which has $d_{HEX} = \frac{0.77}{r^2}$ and square grid which has $d_{SQR} = \frac{1}{r^2}$ [12]. In triangular grids, the number of nodes in a $D \times D$ square area is $\frac{2D}{\sqrt{3}r} . \frac{D}{r} \approx 1.155\frac{D^2}{r^2}$ and hence the density $d_{TRI} = \frac{1.155}{r^2}$.

## 3. Related Work

Existing approaches to create AMRoNet are mostly based on *mobile routers making a chain*. In [4, 13] the authors present different strategies such as Manhattan-Hopper, Hopper, Chase explorer and Go-to-The-Middle for maintaining the connectivity of an explorer with a base station. In [20], depending on whether the knowledge of the agent's trajectory is available or not, the trajectories for the routers are estimated.

The multi-robot spreading algorithms [7, 11, 19] could also be used for AMRoNet robots. In these algorithms, mobile robots spread out based on local rules. If the routers also move out of the base stations pro-actively and spread in the environment, using these algorithms they can form the AMRoNet for agents' communication.

Existing algorithms for multi-robot spreading algorithms is mainly based virtual forces. The most popular force based algorithms use artificial potential field-based forces for coverage maximization. It has been first proposed in [11], and later an extension of this approach to assure K-connectivity has been presented in in [19]. An attractive force $F_{connect}$, refrains the node degree getting too low by making them attract when the node degree becomes critical $(\leq K)$ and a repulsive force $F_{cover}$ keeps it away from obstacles and other robots. The repulsive force $F_{cover}$ and attractive force $F_{connect}$ exerted on node $i$ by its $j^{th}$ neighbour are:

$$F_{cover}(i,j) = \left(\frac{-\mathrm{K}_{\mathrm{cover}}}{distance_{ij}^{2}}\right)\left(\frac{p_i - p_j}{|p_i - p_j|}\right), \text{ where } \left(\frac{p_i - p_j}{|p_i - p_j|}\right) \text{ is the unit vector.} \quad (2)$$

$$F_{connect}(i,j) = \begin{cases} \left(\dfrac{-\mathrm{K}_{\mathrm{cover}}}{(distance_{ij} - r_c)^2}\right)\left(\dfrac{p_i - p_j}{|p_i - p_j|}\right), \text{ if critical connection,} \\ 0 \; otherwise. \end{cases} \quad (3)$$

where $p_i$ and $p_j$ are the position of nodes $i$ and $j$ respectively, and $K_{cover}$ and $K_{degree}$ are the force constants.

Inspired by the equilibrium of molecules, a distributed self-spreading algorithm DSSA has been presented in [7], where the force exerted on node $i$ by its $j^{th}$ neighbour is calculated as:

$$F(i,j) = \frac{D}{\mu^2}\left(r_c - |p_i - p_j|\right)\frac{p_j - p_i}{|p_j - p_i|}, \quad (4)$$

where $|p_i - p_j|$ is the distance, D is the current local density and $\mu^2$ is the expected average density which is $\frac{N.\pi r_s^2}{A}$.

In [24], three Voronoi-based multi-robot spreading algorithms VEC, VOR and Minimax have been proposed. The algorithms uses the structure of Voronoi diagrams for finding coverage holes and minimizes them by relocating the robots. A behavior-based approach for multi-robot spreading has been proposed in [15, 17, 18, 23]. They focus on developing a set of simple local behaviours for exploring unknown terrains.

A virtual force algorithm to enhance the coverage of initial randomly deployed sensors has been presented in [22]. An incremental greedy deployment algorithm is presented in [10] where nodes are deployed sequentially by making use of the information of previously deployed nodes. A robot carrying multiple nodes and deploying them in an unexplored area with the assistance of previously deployed node has been presented in [2]. In [25], a bidding protocol for hybrid sensor networks where the static nodes find coverage holes by constructing Voronoi diagram and bid the mobile nodes to move to the holes is introduced. Maintaining the connectivity of a group of robots while rendezvous, flocking, formation control etc. by controlling their motion pattern has been addressed in [5], [21], [14].

## 4. AD-HOC MOBILE ROBOTIC NETWORKS

Existing approaches to create AMRoNet presented in Section 3 maintain connectivity of the agents, if the routers move as fast as the agents. However, this assumption is not valid in our case as the routers used to create AMRoNet are very simple robots and their speed is usually very small compared to the speed of the agents. The *chain based* approach needs routers that can move faster than the agents [4, 13] and [20] needs twice the speed of the agent, to keep the chain connected. Existing chain based approaches cannot support connectivity of multiple exploring agents. Hence they are not useful in our scenario.

The proactive spreading using *multi-robot spreading* algorithm also needs router moving as fast as the agents to keep them connected. Using simple routers that are slower than the agents, the multi-robot spreading algorithms based approaches work only if the deployment phase is finished prior to the exploration of the agents. However, existing approaches [7, 11, 19, 24] are meant for maximizing coverage with respect to sensing area and not with the communication area. As communication radius is generally larger compared to the sensing radius, an approach that maximizes communication area coverage is interesting.

In this paper we focus on designing an algorithm that uses only *local rules* for AMRoNet creation. Our basic idea is to deploy robots greedily into the environment to those locations that maximize the local coverage and connectivity. We are interested in using the wireless signal intensity information for determining the deployment locations.

We specify two algorithms for deploying routers 1) Agent-assisted router deployment 2) Self-spreading. The *agent-assisted router deployment* algorithm is used in scenarios such as urban search and rescue, where the proactive pre-deployment is not feasible due to the limited speed of the routers compared to the speed of the agents performing search and rescue operations. In scenarios such as exploration of unknown terrains where the proactive pre-deployment is feasible, the self-spreading algorithm is used.

## 4.1. Agent-assisted router deployment algorithm

In the agent-assisted router deployment, the agent carries the routers during the exploration. They release new routers into the area maximizing local communication area coverage. Such an approach is feasible; as our robots are very small [8] and the agents can carry several robots during their exploration.

Let the $N_a$ agents begin their exploration from $n$ base stations. Each base station has a unique id and one reference node which acts as a base station server for all communication. The base station $i$, for all $i \leq n$, is denoted as $BS_i$ and its reference node as $R_i$. We set the *status* of $R_i$ and the agents moving out of $BS_i$ to $i$. Routers are denoted as $R_{ij}$ and agents as $A_{ij}$, where $i$ is their status and $j$ indicates their unique id. The agents explore the area based on their own navigational algorithm. Figure 3 shows a schematic representation with two base stations and two agents (one agent per base station) exploring an open area.

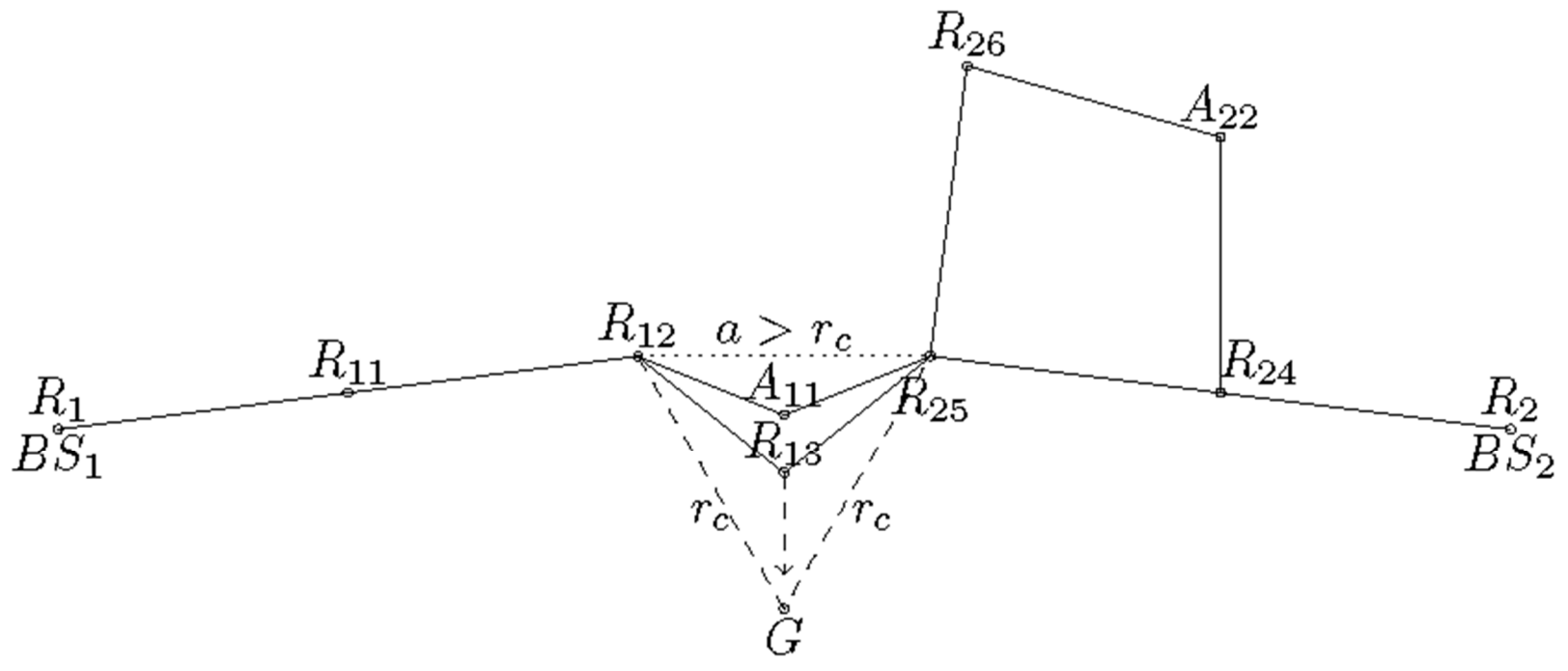

Figure 3: Schematic representation of agent-assisted router deployment in an open region

Initially an agent $A_{ij}$ has wireless links to $R_i$ and other agents $A_{ik}$ for any $k \leq N_a$. As the link between $A_{ij}$ and $R_i$ is initialized, $A_{ij}$ asks $R_i$ about its position and stores this information. If $A_{ij}$ is about to lose its connection to $R_i$, it places a new router with its status set to $i$ and position set to $A_{ij}$'s current position. The new router $R_{ik}$, for any $k \leq N_r$, is placed very close to the current location of $A_{ij}$ in the direction towards $R_i$. Agents use the position information of $R_i$ for estimating the direction. This ensures that $R_{ik}$ released is always connected to $R_i$. $R_{ik}$ becomes the new reference for $A_{ij}$ and for all other agents within $R_{ik}$'s communication range. During the navigation, $A_{ij}$ may move out of $R_{ik}$'s communication range and enter the range of a router $R_{pq}$ for any $p \leq n$ and $q \leq N_r$ that has already been deployed. In this case $R_{pq}$ becomes $A_{ij}$'s current reference. $A_{ij}$ asks $R_{pq}$ for its status and updates its status to $p$ and becomes $A_{pj}$. The agent repeats the placement steps when it is about to lose its connection to its current reference. If an agent has wireless links to many reference robots, any one of them acts as the agent's current reference. The agent releases a new router only when it loses connection to the last reference node in its communication range. We call this placement strategy as *greedy deployment*.

The greedy agent-assisted router deployment builds a graph $G$ with the nodes at the base stations and with routers released during agents' exploration as its vertices. Agents exploring from one base station form a connected component, denoted as $CC$, of $G$. However, such $CC$s created from multiple base stations are not connected. When an agent $A_{xy}$ enters into the range of $R_{pq}$ from the current reference $R_{ij}$, for $i \neq p$ and $i=x$, $CC_i$ and $CC_p$ are temporarily connected. During the navigation, if $A_{xy}$ loses it connection to $R_{ij}$ but still has connection to $R_{pq}$, $A_{ij}$ does not place another router, as it has $R_{pq}$ as its current reference. In this case, $A_{xy}$ loses connection to its previous base station $BS_i$ and $CC_i$ and $CC_p$ get disconnected again.

To solve the disconnection problem, in such situations we adopt another deployment strategy called *triangular deployment*. In triangular deployment, when an agent $A_{xy}$ encounters $R_{pq}$ with from the current reference $R_{ij}$, for $i \neq p$ and $i = x$, it releases a new router $R_{ik}$, for any $k \leq N_r$, and this router moves to a point that keeps $R_{ij}$ and $R_{pq}$ connected and maximizes the local coverage. The goal point of the new router for maximizing the local coverage can be calculated as follows: If $a$ is the distance between $R_{ij}$ and $R_{pq}$, the goal point lies at a distance $d = \sqrt{{r_c}^2 - \left(\frac{a}{2}\right)^2}$ from the midpoint of the line joining $R_{ij}$ and $R_{pq}$ on the same side of the agent as shown in Figure 3 . During the *goto goal* behavior, if the new router encounters an obstacle that cannot be avoided in few steps, it stops navigating to the goal location, as the obstacle could be too large to overcome without disconnecting $R_{ij}$ and $R_{pq}$.

## 4.2. Self-spreading algorithm

In the self-spreading algorithm the routers perform *random-walk* with an obstacle avoidance algorithm. On reaching the points that maximizes local communication area coverage, they stop navigation and become references for others to spread further. The self-spreading algorithm works similar to the agent-assisted router deployment algorithm. The main difference lies in the mode of deployment: In agent-assisted router deployment agents deploy routers whereas in self-spreading the routers themselves navigate and deploy.

We use the same notations given in Section 2.1. for explaining the self-spreading algorithm. Let the routers are initially located in $n$ base stations. Each base station $BS_i$ has one reference node $R_i$, with its status set to $BS_i$, acting as a base station server for all communication. Routers are denoted as $R_{ij}$, where $i$ is their status and $j$ indicates their unique id. Initially the status of the routers is set to -1 and their state to *explore*. Routers with the state *explore* always perform random-walk.

All routers in one $BS_i$ has wireless links to their reference $R_i$. If a router moving out of the base station is about to lose its connection with its reference, it changes its state to *reference* and stops navigation. It sets its status to the status of its previous reference. The new reference $R_{ij}$ becomes the reference for all other exploring routers within the communication radius of $R_{ij}$. The exploring routers use the new reference to navigate further. If an exploring router has wireless links to many reference robots, it chooses any one of them as its current reference. The exploring router changes its state to *reference* and stops navigation only when it loses connection to the last reference in its communication range. This is the *greedy deployment* phase of the self-spreading algorithm.

Routers exploring from one base station form a connected component *CC* and such *CC*s created from multiple base stations are not connected. To solve the disconnection problem, we adopt the same *triangular deployment* strategy uses in Section 2.1. In the triangular deployment, when a router $R_{xy}$ with current reference $R_{ij}$ encounters a news reference $R_{pq}$ with $i \neq p$, it changes its status to $i$ and state to *triangle* and moves to the goal point that keeps $R_{ij}$ and $R_{pq}$ connected and maximizes the local coverage. The goal point is calculated in the same way as in Section 2.1. During the *goto goal* behavior, if it encounters an obstacle that cannot be avoided in few steps, it stops navigation, as the obstacle could be too large to overcome without disconnecting $R_{ij}$ and $R_{pq}$.

### 4.3. Optimization of triangular placement

To optimize the number of robots used during the triangular deployment, we propose two strategies. The first one needs global communication and the second one needs only local communication.

In the *global strategy*, when an agent $A_{xy}$ in the agent assisted algorithm or a router $R_{xy}$ in the self-spreading algorithm entering into the range of $R_{pq}$ from the current reference $R_{ij}$ with $i \neq p$ for triangular deployment, it first checks with $R_{ij}$ and $R_{pq}$ if $CC_i$ and $CC_p$ are already connected. If not, it performs the triangular deployment and connects $CC_i$ and $CC_p$. The router connecting $CC_i$ and $CC_p$ sends a message to all references connected to it either directly or by multi-hop networking informing the new connected components. All these references update the information about the connected components in *G*.

In the *local* strategy, the router deployed sets the references $R_{ij}$, $R_{pq}$ and itself as *disabled* for further triangular deployment. When a router $R_{xy}$ or an agent $A_{xy}$ entering into the range of $R_{pq}$ from the current reference $R_{ij}$ with $i \neq p$, it checks if both $R_{ij}$ and $R_{pq}$ has already been disabled from triangular deployment. This ensures that $CC_i$ and $CC_p$ always get connected and prevents redundant deployment at the locations of triangular placements.

## 5. Experimental Evaluation

We evaluate the proposed agent-assisted router deployment and self-spreading algorithms using a simulation based empirical analysis. We use the open source Player-Stage [6] robotic platform for our experiments. Player abstracts the robot hardware details from the client algorithms and allows them to access it using pre-defined interfaces. Stage simulator provides a virtual world of simulated robots that can be controlled by the client programs via player as if they were real hardware. Thus Stage minimizes the difference between real and simulated robots. The control programs connect to Player server over a TCP socket and read/write data from/to the real or Stage-simulated robots.

### 5.1. Agent-assisted router deployment

We evaluate the agent-assisted router deployment algorithm in a square area of size $32m \times 32m$, which maps the floor plan of our institute as shown in Figure 4. The agents are modeled as Pioneer2dx robots, routers as Bebot robots and base stations' reference robots as Amigobot robots. All robots are equipped with WiFi modules for communication. The base station robots are located at the corner of the simulation environment and are immobile. The scenario shown in Figure 4 has 4 base stations and 12 agents (3 per base station). The agents start their exploration from a point very close to the base station robots and are initially connected to them. We have chosen a random exploration strategy for the agents. They detect obstacles using their sonar sensors which have maximum range of 2m and avoid them using the *obstacle avoidance* behavior implemented. The release of a new router by the agent is implemented by moving a router located outside the simulation environment to its placement point by the simulator.

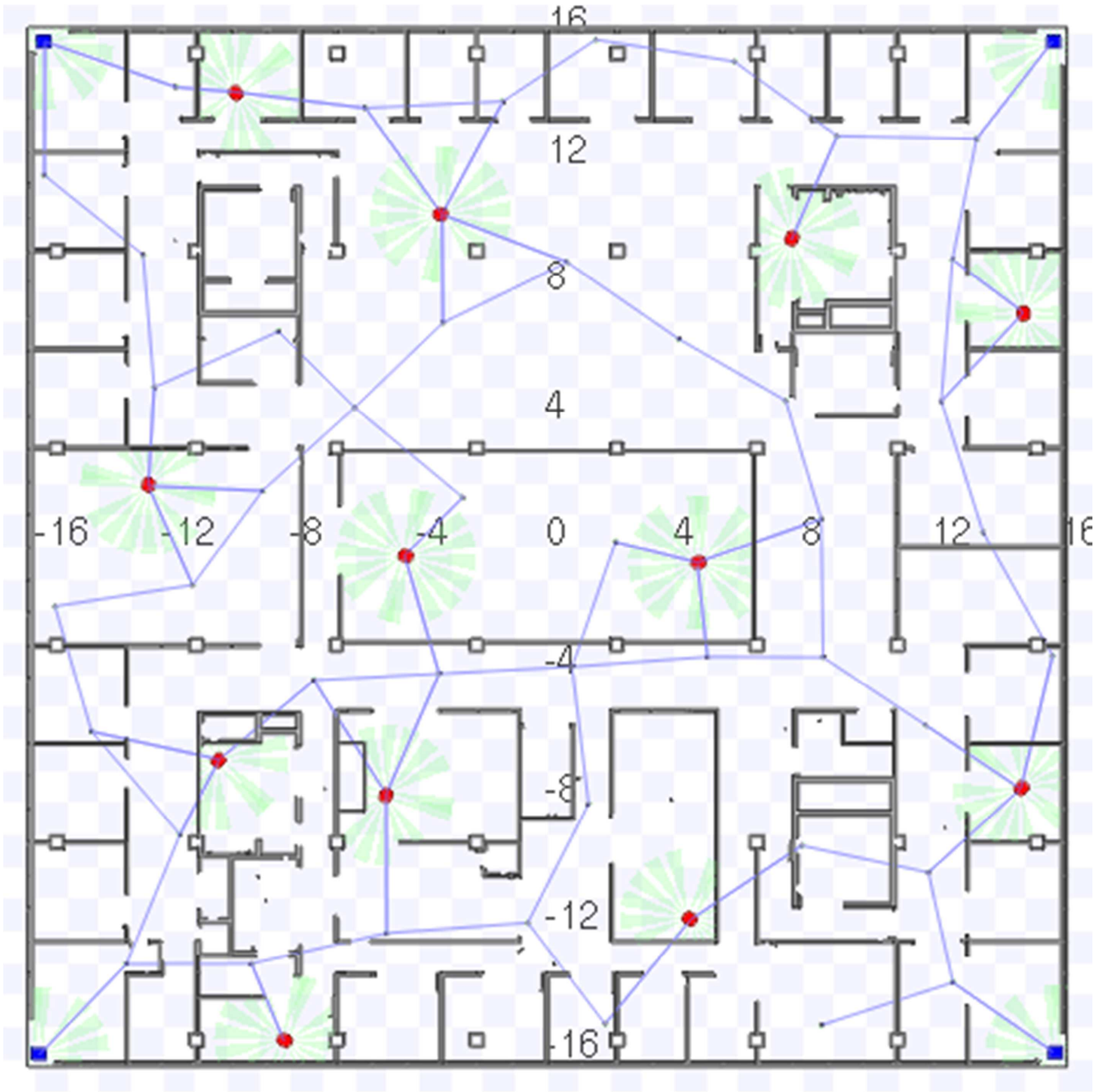

Figure 4: An example scenario with 12 agents and 4 base stations

Routers released during the triangular placement use the *goto* behavior to navigate towards the goal points. They avoid collisions using their IR sensors which have maximum range of 14cm.

### 5.1.1. Performance of agent assisted router deployment algorithm

To analyze the performance of the agent-assisted router deployment algorithm, we vary parameters such as $r_c$ and $N_a$. Figure 5 shows the result of the algorithm, when $r_c$ is varied from 4 to 10 in a square area of size $32m\times32m$. The graph plot with label *ARD* shows the average number of routers (including the reference robot in the base station) deployed to cover the entire region, when all agents begin their exploration from one base station. Here, $N_a$ is varied from 1 to 4. For each $N_a$, the simulation is repeated 5 times and the agents are assigned different start locations. So the graph plot with label *ARD* given in Figure 5 is the average of 20 simulations with confidence interval at 95%.

To compare the performance of the algorithm, we calculate the number of robots required, by the static placement strategies of the commonly used regular patterns such as r-strip tile, hexagonal grid, square grid and triangular grid. The estimated number of robots required to cover the area can be calculated using the spatial density of the patterns, i.e. $d_{STR} = \frac{0.536}{r^2}$, $d_{HEX} = \frac{0.77}{r^2}$, $d_{SQR} = \frac{1}{r^2}$ and $d_{TRI} = \frac{1.155}{r^2}$. Since the area is bounded, the minimum number of robots actually required to cover the entire region is often higher than the estimated values. This is clearly visible in the example figures, Figure 6(a) and Figure 6(b), where the estimated number of robots needed for the r-strip $RSTR_{est}$ is 35 and the hexagonal grid $HEX_{est}$ is 50, but the minimum required number for r-strip tile $RSTR_{min}$ is 44 and the hexagonal grid $HEX_{min}$ is 55. The figures also show that there are still uncovered areas, e.g. the location of the robots highlighted with circles. We cannot place additional routers to cover these areas, as they would be placed outside the specified area according to the regular placement pattern.

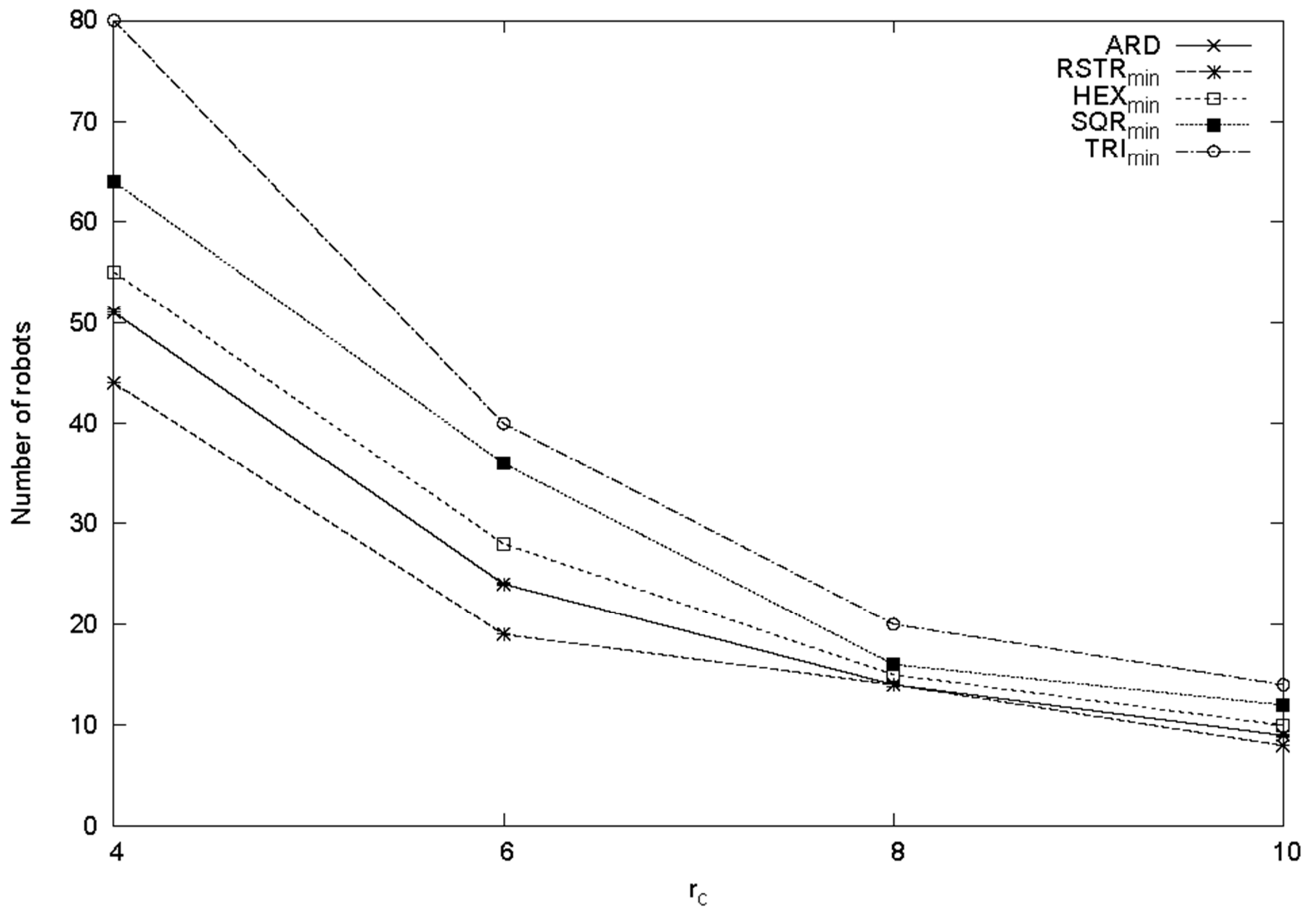

Figure 5: Comparison of performance of router deployment algorithms

Figure 5 also shows the plot of the minimum required values for r-strip tile $RSTR_{min}$, hexagonal grid $HEX_{min}$, square grid $SQR_{min}$ and triangular grid $TRI_{min}$ in the specified square area, when $r_c$ is varied from 4 to 10. It shows that the proposed algorithm is better than $TRI_{min}$, $SQR_{min}$ and $HEX_{min}$ placement strategies. The number of robots needed by the proposed algorithm is close to the $RSTR_{min}$ values which are the actual optimal values.

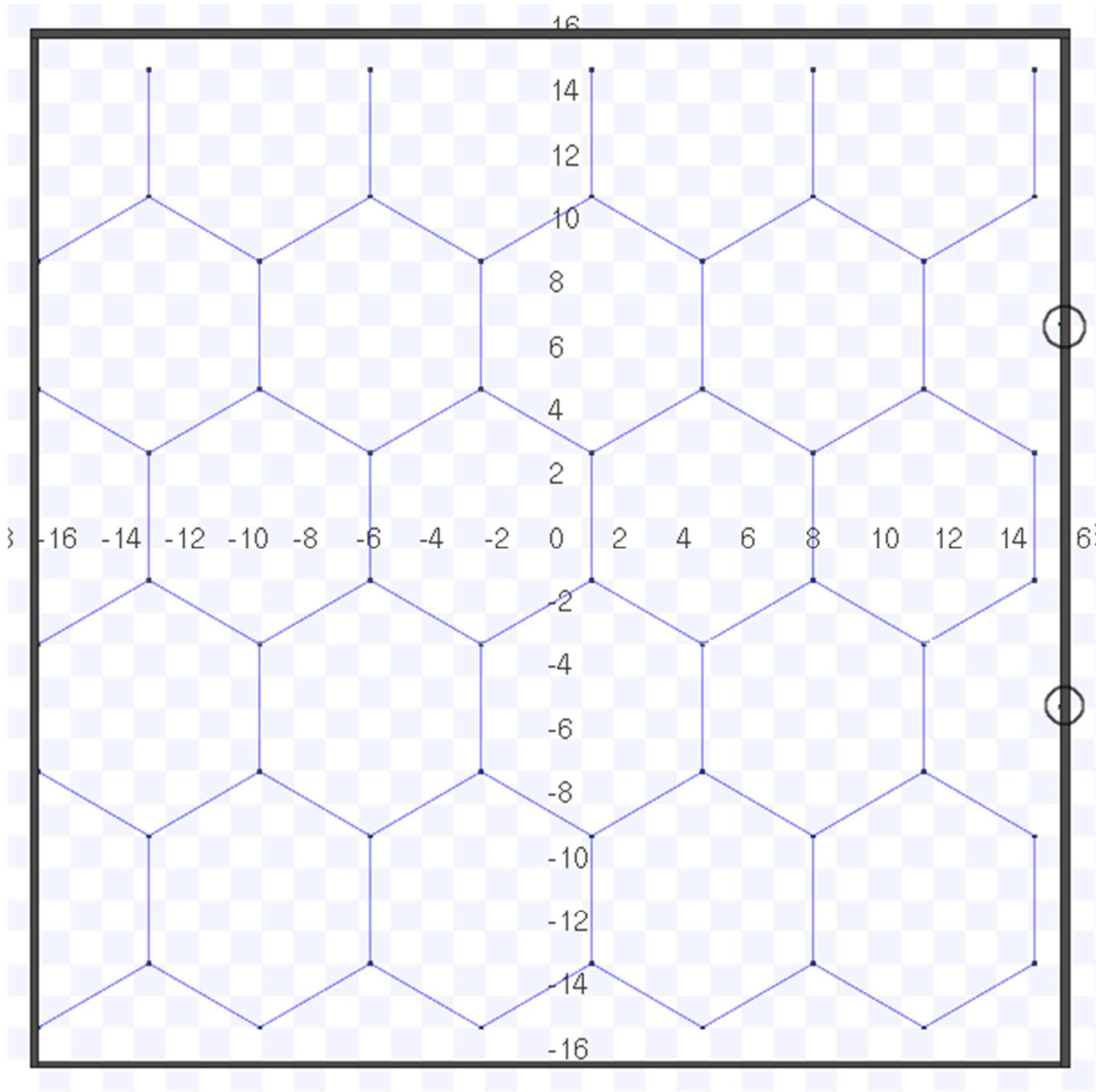

(a) R-strip tile based topology

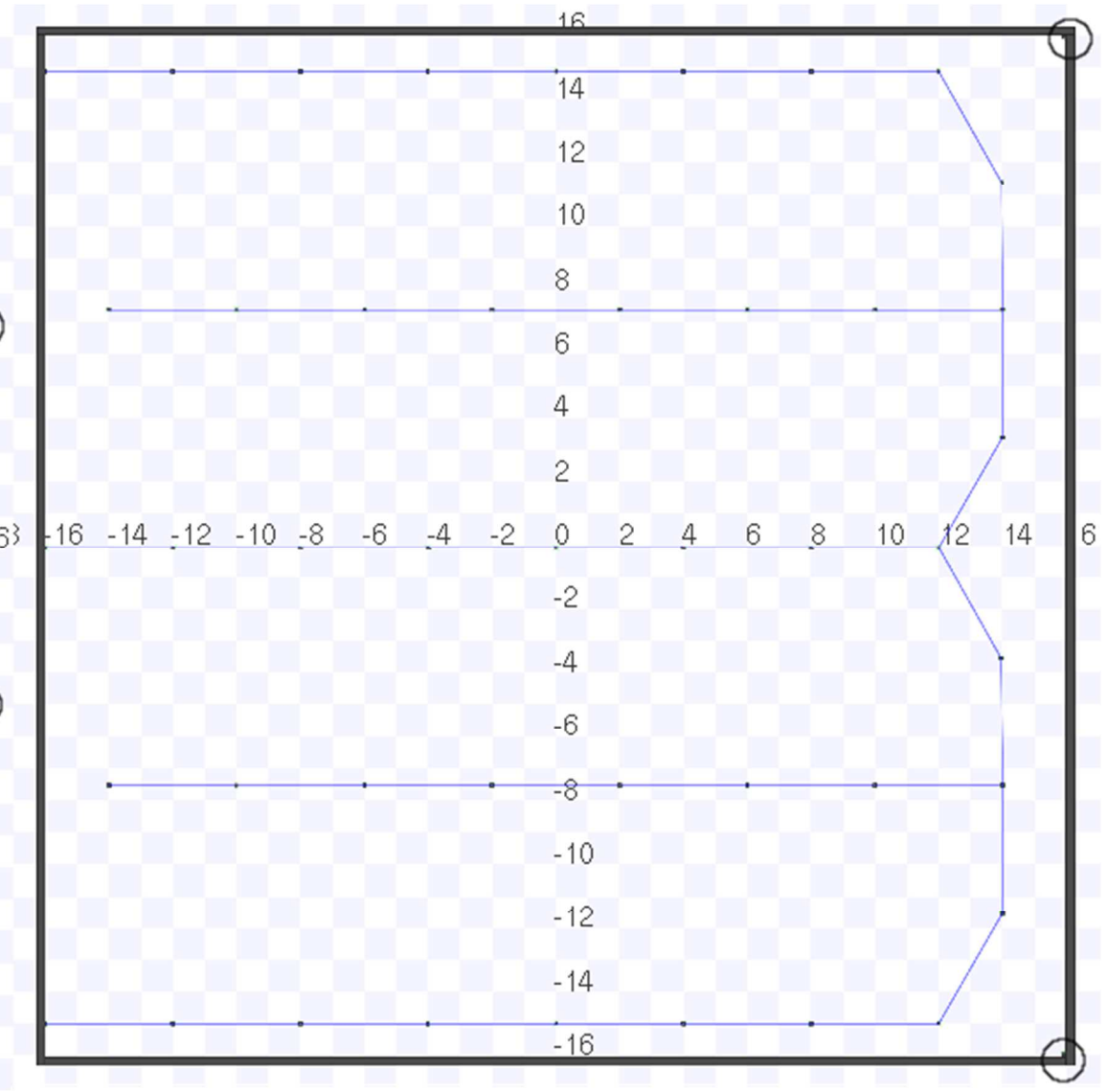

(b) Hexagonal-grid based topology

Figure 6: Static placement of regular pattern

### 5.1.2 Effect of number of agents and base stations

To analyze the effect of number of agents and base stations on the agent-assisted router deployment algorithm, we now vary number of agents per base station $N_{apbs}$ and the number of base stations $n$, for a fixed $r_c$. Figure 7 shows the average number of robots (including the base station robots) needed to cover the square area of size $32m \times 32m$ for $N_{apbs}$ = 1, 2and 3, when $n$ is varied from 1 to 4.

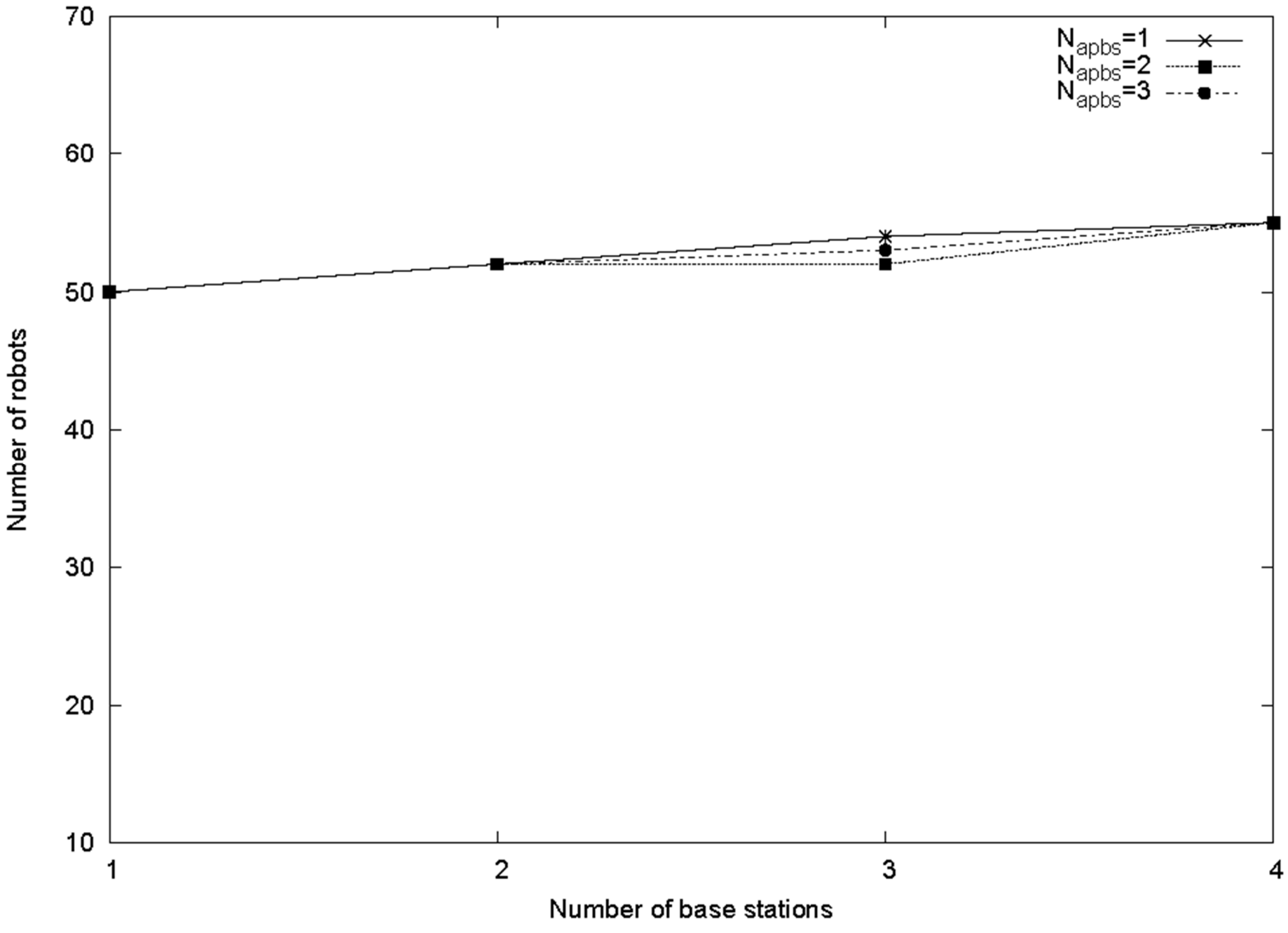

Figure 7: Effect of number of agents and base stations on the performance

Increasing the number of agents without increasing *n* do not affect the performance, as the deployments performed by the agents are based on the local rules which are in turn based only on losing or establishing connection with other routers and not with other agents. Hence the number of routers deployed is independent of the number of agents. The data points for a particular *n* shown in Figure 7 with different $N_{apbs}$ confirm this.

Increasing the number of base stations may result in more triangular deployments. The total area covered by three robots in a triangular deployment is usually lesser than the total area covered by them in an optimal deployment. The largest overlap in a triangular deployment occurs when two references are separated by a distance slightly greater than $r_c$. However, such deployments do not increase the number of routers considerably. Even the greedy deployment may produce similar less optimal overlapping regions, e.g. when an agent connected to two references move out of the communication radius of both references simultaneously.

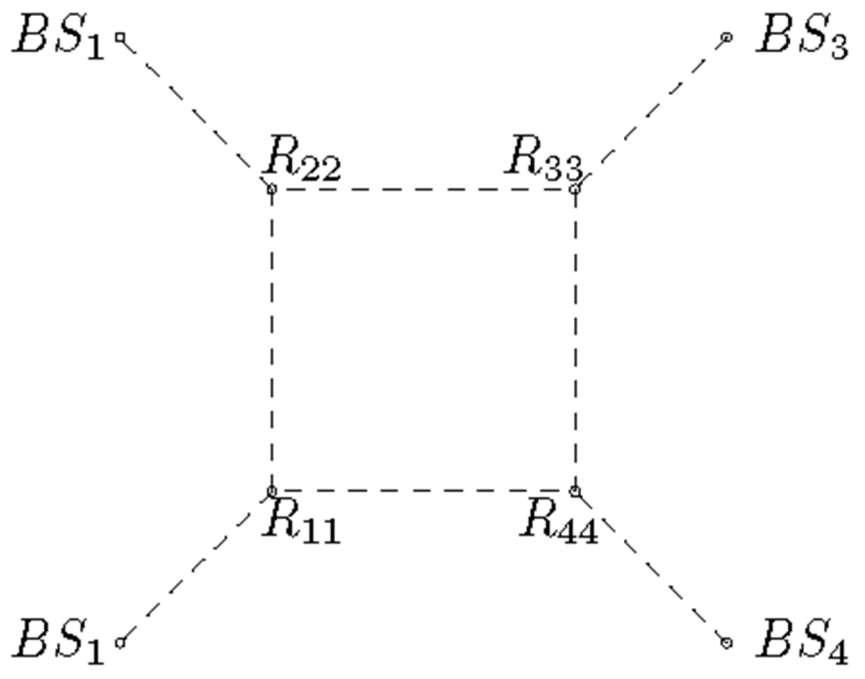

Figure 8: Redundant router deployment during local triangular deployment

Figure 8 shows a scenario where three routers are released during the triangular deployment. Actually at most 2 routers are needed to make the four chains connected. Such redundant deployment increases with the number of base stations. We could add more local rules to make the increase bounded, but this is not actually needed as the agents move independently (in our experiments, they move randomly) and the structures similar to the one shown in Figure 8 occur very rarely. The graph plots for $N_{apbs} = 1$ and $N_{apbs} = 2$ depicted in Figure 7 also show that the total number deployed is more or less the same for different base station counts.

## 5.2. Self-spreading algorithm

We evaluate the self-spreading algorithm in a square area of size $6m\times6m$ as shown in Figure 9. We use the Bebots robots as the routers. They are equipped with WiFi modules for communication. We set the communication radius $r_c$ to 1m. Initially, 30 Bebots are deployed randomly at a base station, which is a squared region of size $1m\times1m$ at the centre of the region. The base station has one reference robot, which is located at the origin *(0,0)*. We use the Bebot robot with id 1 to indicate the reference robot. The reference robot is immobile. All robots in the base station are initially connected to the reference robot. The routers start their random walk from the base station and spread according the self-spreading algorithm described in Section 4.2. Routers avoid collisions using their IR sensors which have maximum range of 14cm.

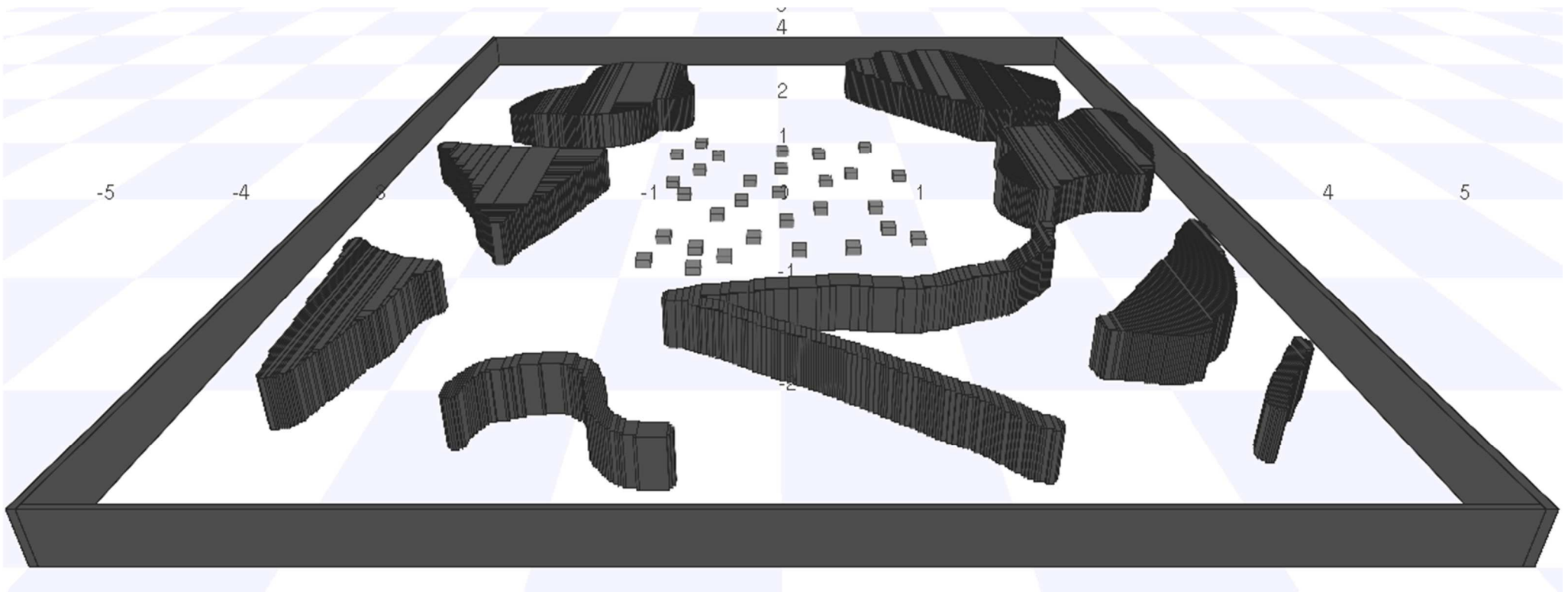

Figure 9: An example scenario with obstacles

### 5.2.1. Performance of self-spreading algorithm

*Experiment 1 - Square map with no obstacles:* The first experiment to compare the performance of self-spreading is conducted on a square field of size $6m \times 6m$ with no obstacles in the area. The robots spread according to the following three algorithms DSSA, potential field and self-spreading. For the potential field algorithm, we find the constants $K_{cover}$ and $K_{degree}$ of equation 1 and 2 exactly as mentioned in original paper [19] and assign the same values 0.25 and 0.8 for the damping and safety factors as used in the paper. For DSSA, the expected density μ and the local density *D* are calculated exactly as mentioned in the paper [7]. The sensing radius used by these algorithms to find the constants was set to *0.5*.

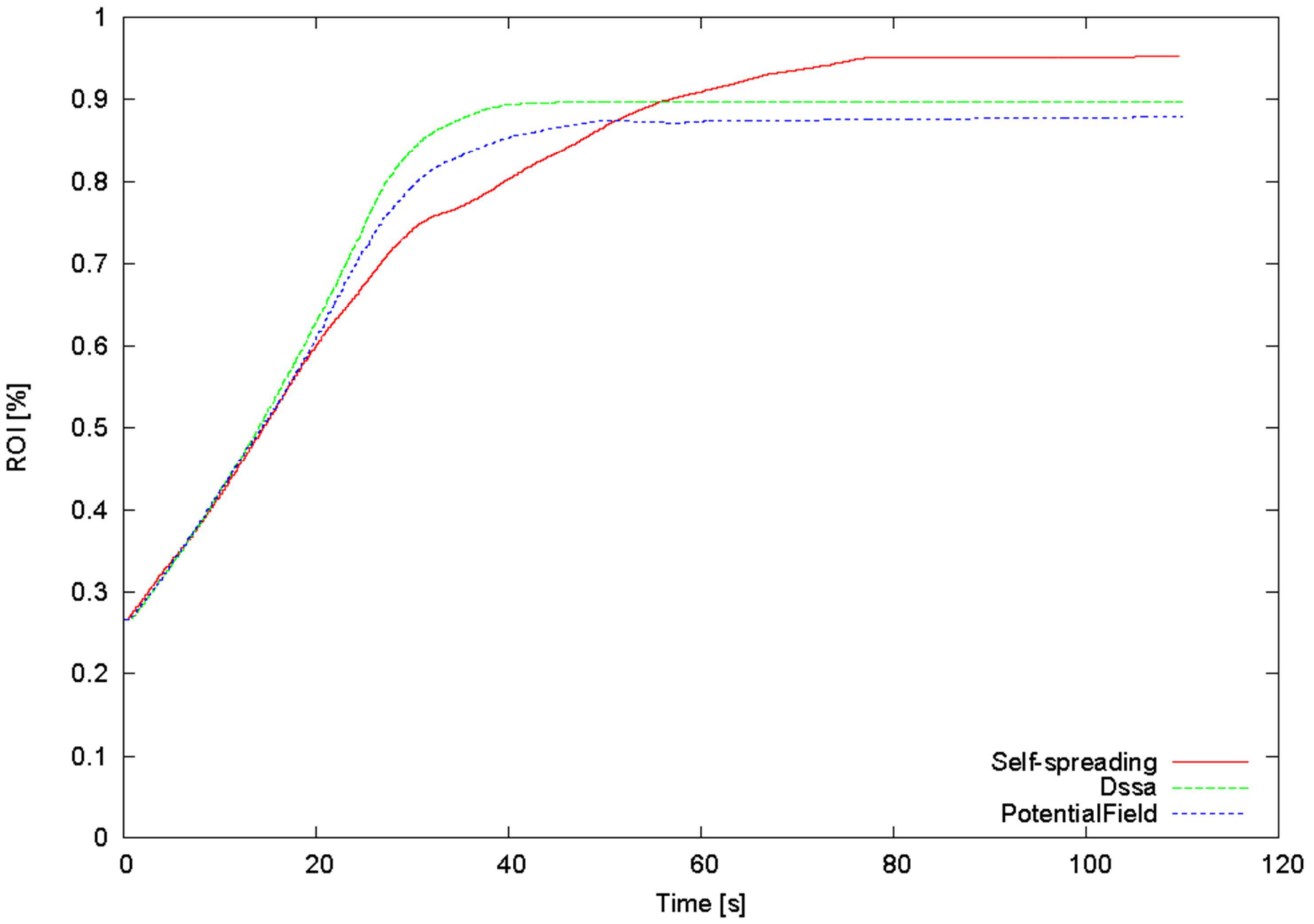

Figure 10: Comparison of performance on square map without obstacles

Figure 10 shows the performance of the three algorithms averaged over 50 independent runs with different random initial configurations. It shows that the self-spreading algorithm performs better than state of the art algorithms by covering more area with time. Initially the coverage of the self-spreading is slightly lower than other algorithms. This is due to the fact that the force vector used by the state-of-the-art to push robots apart very strongly when the robots are very close, whereas in random walking strategy they move with almost same speed (exception: encountering obstacles) irrespective of the distance of separation between the robots.

*Experiment 2 - Square map with obstacles:* In this experiment we test the algorithms on the same square field of size *6mx6m*, but with obstacles as shown in Figure 9. Forces from the obstacles are also added to the force calculation of DSSA and potential field algorithms.

Figure 11 shows the performance of the DSSA, potential field and self-spreading algorithm averaged over 50 independent runs with different random initial configurations. It shows that the performance of the self-spreading algorithm is better than the state-of-the-art algorithms. The performance of DSSA and potential field degrade with presence of obstacles in the environment, whereas self-spreading achieves almost the same final coverage as in the experiments without obstacles.

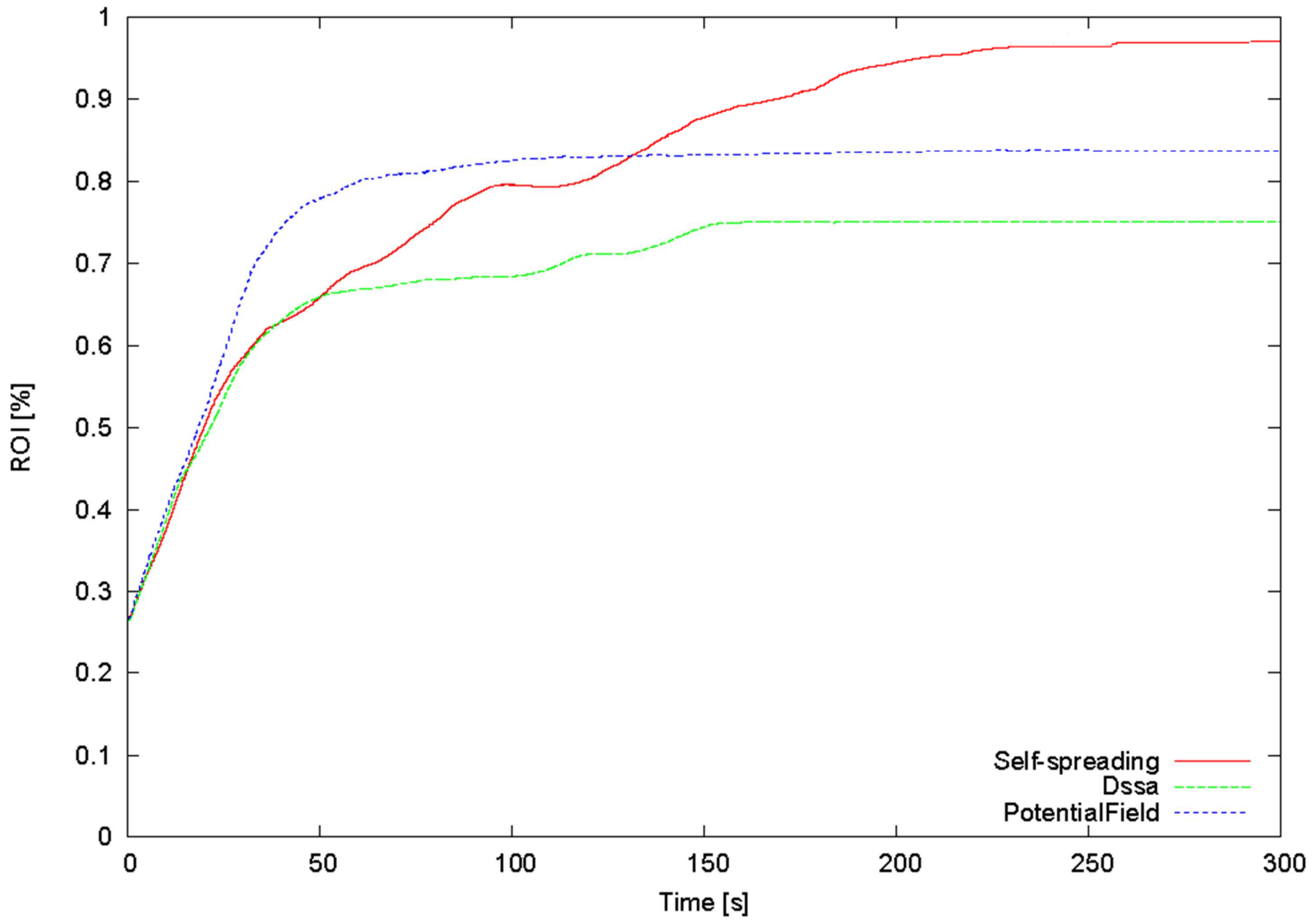

Figure 11: Comparison of performance on square map with obstacles

## 6. Performance Analysis

### 6.1. Analysis of agent-assisted router deployment algorithm

From the evaluation of the proposed agent-assisted router deployment algorithm, we see that it performs better than all other regular pattern based static placements except the r-strip tile in 2-D. Let us now look at a localized agent-assisted r-strip tile creation algorithm that does not restrict agent movements or causes disconnections. A straight forward extension of the agent-assisted router deployment algorithm for r-strip creation is: Agents release routers as per the greedy deployment strategy and the routers move to the goal points that create r-strip tiles locally.

During this localized r-strip creation the following problems arises: The routers released move to their goal point very slowly compared to the agent speed. If the agents use these moving routers as their references, to prevent disconnections they may have to release new routers before their current references reach their goal points. Hence more routers than the static optimal r-strip tiles are needed for this localized solution. Another problem is the presence of obstacles which prevents the routers from reaching the ideal optimal goal point. A third problem occurs when we have multiple base stations. The pattern created from one base station may not be aligned with the other from another base station. This also affects the optimality of the localized r-strip creation algorithm. These problems are not specific to the localized r-strip tile creation algorithm. The localized algorithms for creating regular patterns like hexagon, square or triangular grids also suffers these problems. Another problem that is specific to r-strip tile creation algorithm is: Non-optimal placement of the vertical strip that is needed to connect different horizontal r-strips. In the ideal case, it needs only one router to connect two horizontal strips. However, if the agents move in a adversarial manner, it needs one router per every second router in the horizontal strip.

The localized r-strip creation without restricting agent movements or causing disconnection is not an optimal solution due to the problems mentioned above. Hence the actual number of robots needed for localized r-strip creation is much more than the estimated static r-strip tile value. Figure 5 shows that the agent assisted router deployment algorithm's performance is quite close to the actual static r-strip tile $RSTR_{min}$ value. Hence it is one of the best localized approaches to create an AMRoNet.

If we calculate the estimated number of robots needed for the hexagonal grid $HEX_{est}$ in the specified square region for different $r_c$ values, we observe that they are very close to the average number of routers used by the proposed algorithm. Hence we could use the equation $ARD_{est} = \frac{0.77}{r^2} * A$ to get an approximate estimate of the total number of routers needed to cover a given area $A$. This helps the agents in making an estimate on the numbers routers they need to carry, before beginning their exploration.

## 6.2. Analysis of self-spreading algorithm

From the evaluation of the proposed self-spreading algorithm, we see that it works better than the force based approaches such as potential field and DSSA by covering more regions with time. The main limitation of the force-based approaches is that, they are meant for maximizing the sensing area coverage. To make them comparable with our self-spreading algorithm, we used a virtual sensing radius $r_s = 0.5r_c$ such that when they try to maximize the sensing area coverage (i.e. $2\, r_s$), it is equivalent to maximizing the communication area coverage. Moreover, the force-based approaches work only when the routers have precise location information about their neighbors. Usually the $r_s << r_c$ and estimating neighbors position beyond the sensing range is non-trivial, especially when the absolute positioning devices such as GPS [9] are not available due to the limitations imposed by the application scenarios or the limitations of the swarm robots. Even small errors in estimating the orientation of the neighbours lead to incorrect force calculation and degraded performance. So they are not useful for maximizing communication area coverage in realistic environments. Our approach could be easily extended to make it make it work without any location information.

In some experiments, the initial coverage of the self-spreading is slightly lower than other algorithms. We have discussed in Section 5.2.1. that it is due to constant speed used in random-walk irrespective of the distance of separation between the robots. Since self-spreading follows the random-walk based exploration, at some instances the mobile routers get trapped without knowing where to move to achieve better coverage. They may revisit the covered area during their random-walk. These disadvantages of the simple random-walking strategy used could be solved by adding more local rules to move them to the uncovered regions during their random-walk.

## 6.3. Merits of agent-assisted router deployment and self-spreading algorithms

From our experiments, we found that the proposed algorithms performed equally well, irrespective of the presence of the obstacles in the area. The performance of the existing localized state-of-the art algorithms degrades with the presence of obstacles. Our approach even works in area where we do not have any prior model or map of the environment. It could be extended to make it work without any location information, where we need just the link quality estimate provided by the WiFi devices. In such cases, the greedy deployment strategy is performed when the link quality drops below a threshold. Routers deployed during the triangular deployment, move in the direction where the link quality tends to be weak, in order to maximize the coverage area.

## 7. CONCLUSION

We have presented two new localized and distributed algorithms for creating an ad-hoc mobile router network that facilitates communication between the agents without restricting their movements. The first algorithm, *agent-assisted router deployment, is* used in scenarios where a proactive pre-deployment is not feasible due to the limited speed of the routers compared to the speed of the agents and the second one *self-spreading* is used in scenarios where the proactive pre-deployment is feasible. The algorithms have a greedy deployment strategy for releasing new routers effectively into the area and a triangular deployment strategy for connecting different connected components created from different base stations.

Empirical analysis of the agent-assisted router deployment algorithm shows that the number of routers deployed by the algorithm is close to the optimal static r-strip tile values and the analysis of self-spreading algorithm shows that the performance is better than the state-of-the-art force-based multi-robot spreading algorithms. The performance of our algorithms is not affected by the presence of obstacles in the environment whereas it degrades in the state-of-the-art algorithms.

We plan to verify the performance of the proposed algorithm in real life scenarios. The algorithms could be extended to make it work without any location information, using the link quality estimate provided by the WiFi devices. The performance of the algorithm in such cases needs to be validated with more quantitative results. We used a simple random-walking strategy for the routers in the self-spreading algorithm. This could be improved by adding more local rules and make the exploration more efficient.

## REFERENCES

[1] Bai, X., Kumar, S., Xuan, D., Yun, Z., Lai, T.H.: Deploying wireless sensors to achieve both coverage and connectivity. In: Proceedings of the 7th ACM international symposium on Mobile ad hoc networking and computing. pp. 131–142. MobiHoc ’06, ACM, New York, NY, USA (2006)

[2] Batalin, M., Sukhatme, G.: The design and analysis of an efficient local algorithm for coverage and exploration based on sensor network deployment. Robotics, IEEE Transactions on 23(4), 661 –675 (Aug 2007)

[3] Clark, B.N., Colbourn, C.J., Johnson, D.S.: Unit disk graphs. Discrete Mathematics 86(1-3), 165–177 (1990)

[4] Dynia, M., Kutylowski, J., auf der Heide, F.M., Schrieb, J.: Local strategies for maintaining a chain of relay stations between an explorer and a base station. In: Proceedings of the nineteenth annual ACM symposium on Parallel algorithms and architectures. pp. 260–269. SPAA ’07, ACM, New York, NY, USA (2007)

[5] Ganguli, A., Cortes, J., Bullo, F.: Multirobot rendezvous with visibility sensors in nonconvex environments. IEEE Transactions on Robotics 25(2), 340 –352 (April 2009)

[6] Gerkey, B., Vaughan, R.T., Howard, A.: The player/stage project: Tools for multi-robot and distributed sensor systems. In: ICAR'03, Proceedings of the 11th International Conference on Advanced Robotics. pp. 317–323. New York, NY, USA (June 2003)

[7] Heo, N., Varshney, P.: A distributed self spreading algorithm for mobile wireless sensor networks. In: Wireless Communications and Networking, 2003. WCNC 2003. 2003 IEEE. vol. 3, pp. 1597 –1602 (March 2003)

[8] Herbrechtsmeier, S., Witkowski, U., RÃ¼ckert, U.: Bebot: A modular mobile miniature robot platform supporting hardware reconfiguration and multi-standard communication. In: Progress in Robotics, Communications in Computer and Information Science, vol. 44, pp. 346–356. Springer Berlin Heidelberg (2009)

[9] Hofmann-Wellenhof, B., Lichtenegger, H., Collins, J.: Global Positioning System: Theory and Practice. Springer-Verlag, (1997).

[10] Howard, A., Matarić, M.J., Sukhatme, G.S.: An incremental self-deployment algorithm for mobile sensor networks. Autonomous Robots Special Issue on Intelligent Embedded Systems 13(2), 113–126 (2002)

[11] Howard, A., Mataric, M.J., Sukhatme, G.S.: Mobile sensor network deployment using potential fields: A distributed, scalable solution to the area coverage problem. In: DARS: 6th International Symposium on Distributed Autonomous Robotics Systems. pp. 299–308 (June 2002)

[12] Iyengar, R., Kar, K., Banerjee, S.: Low-coordination topologies for redundancy in sensor networks. In: Proceedings of the 6th ACM international symposium on Mobile ad hoc networking and computing. pp. 332–342. MobiHoc '05, ACM, New York, NY, USA (2005)

[13] Kutyowski, J., Meyer auf der Heide, F.: Optimal strategies for maintaining a chain of relays between an explorer and a base camp. Theor. Comput. Sci. 410, 3391–3405 (August 2009)

[14] Lin, Z., Broucke, M., Francis, B.: Local control strategies for groups of mobile autonomous agents. IEEE Transactions on Automatic Control 49(4), 622 – 629 (April 2004)

[15] Ludwig, L., Gini, M.: Robotic swarm dispersion using wireless intensity signals. In: International Symposium on Distributed Autonomous Robotic Systems, Minnesota, USA (July 2006)

[16] Mathews, Emi; Mathew, Ciby: Connectivity of Autonomous Agents Using Ad-hoc Mobile Router Networks. In: Third International Conference on Networks & Communications, LNICST, Bangalore, India, (Jan. 2012)

[17] McLurkin, J., Smith, J.: Distributed algorithms for dispersion in indoor environments using a swarm of autonomous mobile robots. 7th International Symposium on Distributed Autonomous Robotic Systems (DARS) (2004)

[18] Pakanati, A., Gini, M.: Swarm dispersion via potential fields, leader election, and counting hops. In: Proceedings of the Second international conference on Simulation, modeling, and programming for autonomous robots. pp. 485–496. SIMPAR'10, Springer-Verlag (2010)

[19] Poduri, S., Sukhatme, G.S.: Constrained coverage for mobile sensor networks. In: IEEE International Conference on Robotics and Automation. pp. 165–172. New Orleans, LA (May 2004)

[20] Tekdas, O., Wei, Y., Isler, V.: Robotic routers: Algorithms and implementation. Int. J. Rob. Res. 29, 110–126 (January 2010)

[21] Zavlanos, M., Jadbabaie, A., Pappas, G.: Flocking while preserving network connectivity. In: 46th IEEE Conference on Decision and Control. pp. 2919 –2924 (Dec 2007)

[22] Zou, Y., Chakrabarty, K.: Sensor deployment and target localization based on virtual forces. In: INFOCOM 2003. Twenty-Second Annual Joint Conference of the IEEE Computer and Communications. IEEE Societies. vol. 2, pp. 1293 – 1303 (April 2003)

[23] Ugur, E., Turgut, A., Sahin, E.: Dispersion of a swarm of robots based on realistic wireless intensity signals. In: ISCIS 2007. 22nd International symposium on Computer and Information sciences. pp. 1 –6 (November 2007)
[24] Wang, G., Cao, G., La Porta, T.F.: Movement-assisted sensor deployment. IEEE Transactions on Mobile Computing 5, 640–652 (June 2006)
[25] Wang, G.G., Cao, G., Berman, P., La Porta, T.F.: Bidding protocols for deploying mobile sensors. IEEE Transactions on Mobile Computing 6, 563–576 (May 2007)

**Authors**

Emi Mathews is PhD candidate in the Design of Distributed Embedded Systems research group at Heinz Nixdorf Institute, University of Paderborn, Germany. He is a fellowship holder of the Inter-national Graduate School at the University of Paderborn. His research interests are cyber physical systems, swarm robotics, ad-hoc network routing protocols and machine learning. He focuses on building bio-inspired self-organizing systems.

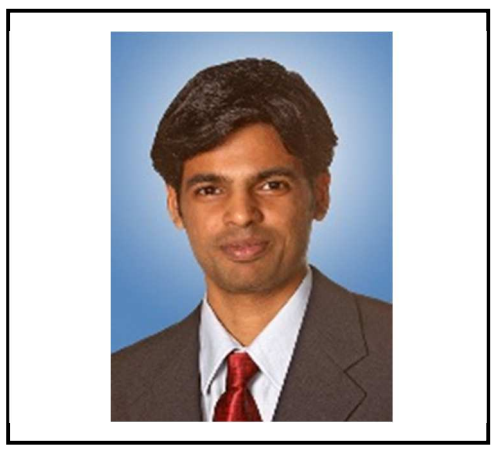

Ciby Mathew is a master student in the Department of Computer Science, University of Paderborn, Germany. He received his bachelor degree in Computer Science from the College Of Engineering Chengannur, India. He specializes in the area of embedded systems. His research interests are swarm robots, object tracking and mobile ad-hoc networks.

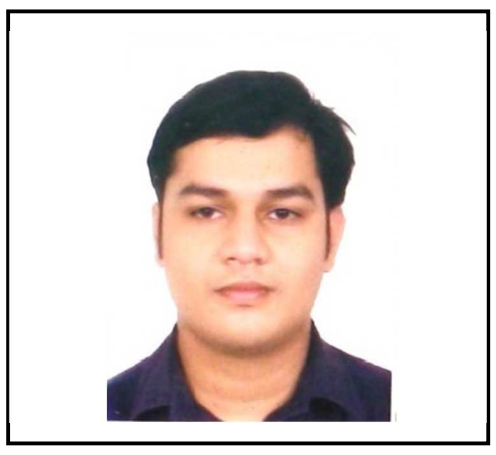